\documentclass[conference]{IEEEtran}
\IEEEoverridecommandlockouts
\usepackage{cite}
\usepackage{amsmath,amssymb,amsfonts}
\usepackage{algorithmic}
\usepackage{graphicx}
\usepackage{textcomp}
\usepackage{xcolor}
\usepackage{hyperref}
\usepackage{multirow}
\usepackage{booktabs}
\usepackage{subcaption}
\usepackage{float}
\usepackage{soul}
\usepackage[export]{adjustbox}

\def\BibTeX{{\rm B\kern-.05em{\sc i\kern-.025em b}\kern-.08em
T\kern-.1667em\lower.7ex\hbox{E}\kern-.125emX}}
\begin{document}

\title{TAME: Attention Mechanism Based Feature Fusion for Generating Explanation Maps of Convolutional Neural Networks
\thanks{This work was supported by the EU Horizon 2020 programme under grant agreement H2020-101021866 CRiTERIA.}
}

\author{\IEEEauthorblockN{Mariano Ntrougkas}
\IEEEauthorblockA{\textit{CERTH-ITI} \\
Thessaloniki, Greece, 57001 \\
ntrougkas@iti.gr}
\and
\IEEEauthorblockN{Nikolaos Gkalelis}
\IEEEauthorblockA{\textit{CERTH-ITI} \\
Thessaloniki, Greece, 57001 \\
gkalelis@iti.gr}
\and
\IEEEauthorblockN{Vasileios Mezaris}
\IEEEauthorblockA{\textit{CERTH-ITI} \\
Thessaloniki, Greece, 57001 \\
bmezaris@iti.gr}
}

\maketitle

\IEEEpubidadjcol

\begin{abstract}
The apparent ``black box'' nature of neural networks is a barrier to adoption in applications where explainability is essential.
This paper presents TAME (Trainable Attention Mechanism for Explanations)\footnote{Source code is made publicly available at: \url{ https://github.com/bmezaris/TAME}}, a method for generating explanation maps with a multi-branch hierarchical attention mechanism.
TAME combines a target model's feature maps from multiple layers using an attention mechanism, transforming them into an explanation map.
TAME can easily be applied to any convolutional neural network (CNN) by streamlining the optimization of the attention mechanism's training method and the selection of target model's feature maps.
After training, explanation maps can be computed in a single forward pass.
We apply TAME to two widely used models, i.e. VGG-16 and ResNet-50, trained on ImageNet and show improvements over previous top-performing methods.
We also provide a comprehensive ablation study comparing the performance of different variations of TAME's architecture.\footnote{\copyright2022 IEEE. Personal use of this material is permitted. Permission from IEEE must be obtained for all other uses, in any current or future media, including reprinting/republishing this material for advertising or promotional purposes, creating new collective works, for resale or redistribution to servers or lists, or reuse of any copyrighted component of this work in other works.}
\end{abstract}


\section{Introduction}

Convolutional neural networks (CNNs) \cite{NIPS2012_c399862d} have achieved exceptional performance in many important visual tasks
such as breast tumor detection \cite{chiao2019detection}, video summarization 
\cite{apostolidis_ICMR_22} and event recognition \cite{Gkalelis_2021_CVPR}.
The trade-off between model performance and explainability, and the end-to-end learning strategy, leads to the development of CNNs that many times act as ``black box'' models that lack transparency \cite{HamonIEEECLM2022}.
This fact makes it difficult to convince users in critical fields, such as healthcare, law, and governance to trust and employ such systems, thus limiting the adoption of AI \cite{amann2020explainability,HamonIEEECLM2022}.
Therefore, it is necessary to develop solutions that address these challenges.

\begin{figure}[tbp]%
\centering
\includegraphics[width=3.5in]{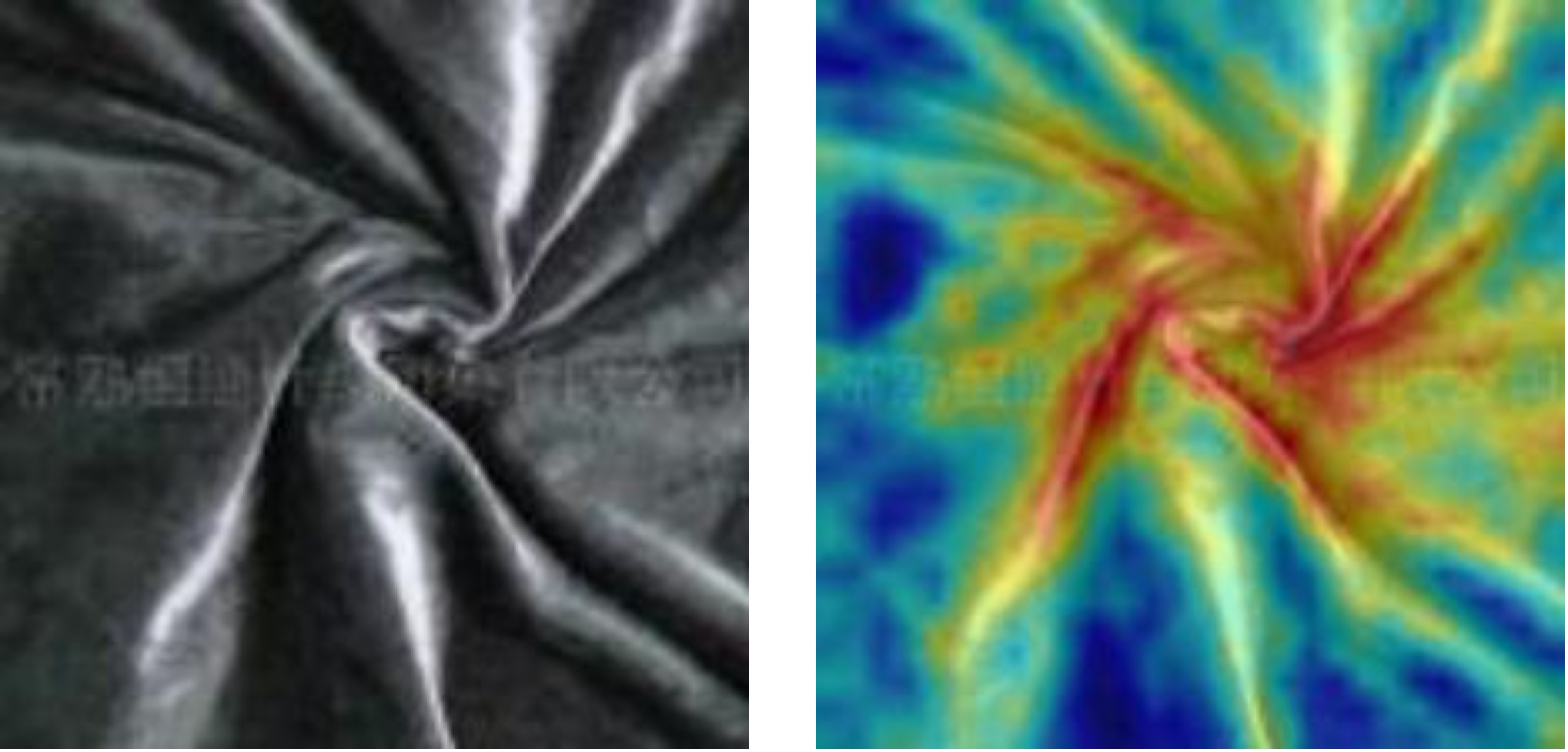}
\caption{An explanation produced by TAME.
The input image belongs to the class "velvet", which cannot be localized.
The produced explanation highlights the salient features of the image explaining the decision of the classifier.
}
\label{fig:XaiVsLoc}
\end{figure}%

Explainable artificial intelligence (XAI) is an active research area in machine learning.
XAI focuses on developing explainable techniques that help users of AI systems to comprehend, trust and more efficiently manage them \cite{arrieta2020explainable,samek2021explaining}.
For the image classification task, a diverse range of post-hoc explanation approaches exist that in a second step take the trained model and try to uncover its decision strategy \cite{samek2021explaining}.
These methods produce an explanation map, highlighting salient input features.
We should note that these methods should not be confused with approaches targeting weakly supervised learning tasks such as weakly supervised object localization or segmentation  \cite{JiangTPAMI2021}, which also generate heatmaps as an intermediate step, and their goal is to locate the region of the target object rather than to explain the classifier's decision (e.g. see the example depicted in Fig. \ref{fig:XaiVsLoc}).

Gradient-based methods \cite{selvaraju2017grad,chattopadhay2018grad} were probably among the first to appear in the XAI domain.
These methods use gradient information to produce explanations, but they are strongly affected by noisy gradients, and the explanations contain high-frequency variations \cite{AdebayoNIPS2018}.
Perturbation-based methods \cite{petsiuk2018rise,wang2020score}, perturb the input and observe changes in the output, thus do not suffer from gradient-based problems as above.
Similarly, response-based methods \cite{sattarzadeh2021explaining,sudhakar2021ada,englebert2022backward} combine a model's intermediate representations, or features, to generate explanations.
However, most methods of the two latter categories described above are computationally expensive because each input requires many forward passes for an accurate explanation map to be produced.

To address the above limitation, L-CAM \cite{Gkartzonika2022} trains an attention mechanism to combine feature maps from the last convolutional layer of a frozen CNN model and produce high quality explanations in one forward pass.
However, L-CAM, by design, uses the feature maps of only the last convolutional layer, and thus, may not be able to adequately capture all the information contained in the CNN model.
To this end, we propose TAME (Trainable Attention Mechanism for Explanations), which exploits intermediate feature maps extracted from multiple layers of any CNN model.
These features are then used to train a multi-branch hierarchical attention architecture for generating class-specific explanation maps in a single forward pass.
We provide a comprehensive evaluation study of the proposed method on ImageNet \cite{ImageNet} using two popular CNN models (VGG-16 \cite{simonyan2014very}, ResNet-50 \cite{he2016deep}) and popular XAI measures \cite{chattopadhay2018grad}, demonstrating that TAME achieves improved explainability performance over other top-performing methods in this domain.

\section{Related Work}

In this section, we briefly survey the state-of-the-art XAI approaches that are mostly related to ours.
For a more comprehensive review the interested reader is referred to \cite{arrieta2020explainable,samek2021explaining}.


Most XAI approaches can be roughly categorized into response-, gradient- and perturbation-based.
Gradient-based methods \cite{selvaraju2017grad,chattopadhay2018grad} compute the gradient of a given input with backpropagation and modify it in various ways to produce an explanation map.
Grad-CAM \cite{selvaraju2017grad}, one of the first in this category, uses global average pooling in the gradients of the target network's logits with respect to the feature maps to compute weights.
The explanation maps are obtained as the weighted combination of feature maps and the computed weights.
Grad-CAM++ \cite{chattopadhay2018grad} similarly uses gradients to generate explanation maps.
These methods suffer the same issues as the gradients they use: neural network gradients can be noisy and suffer from saturation problems for typical activation functions such as ReLU and Sigmoid \cite{AdebayoNIPS2018}.

Perturbation-based methods \cite{petsiuk2018rise,wang2020score} alter the input and produce explanations based on the change in the confidence of the original prediction; thus, avoid problems related with noise gradients.
For instance, RISE \cite{petsiuk2018rise} utilizes Monte Carlo sampling to generate random masks, which are then used to perturb the input image and generate a respective CNN score.
Using the generated scores as weights, the explanation is derived as the weighted combination of the various random masks.
Thus, RISE, as most methods in this category, requires many forward passes through the network to generate an explanation, increasing the inference time considerably.

Finally, response-based methods \cite{sattarzadeh2021explaining,sudhakar2021ada,englebert2022backward,Gkartzonika2022} use feature maps or activations of layers in the inference stage to interpret the decision-making process of a neural network.
One of the earliest methods in this category, CAM \cite{zhou2016learning}, uses the output of the global average pooling layer as weights, and computes the weighted average of the features maps at the final convolutional layer.
CAM requires the existence of such a global average pooling layer, restricting its applicability to only this type of architectures.
SISE \cite{sattarzadeh2021explaining}, and later Ada-SISE \cite{sudhakar2021ada}, aggregate feature maps in a cascading manner to produce explanation maps of any DCNN model.
Similarly, Poly-CAM \cite{englebert2022backward} upscales feature maps to the dimension of the largest spatial dimension feature map and combines them in a cascading manner.
The above methods require many forward passes to produce an explanation. 
L-CAM \cite{Gkartzonika2022} mitigates the above limitation using a learned attention mechanism to compute class-specific explanations in one forward pass.
However, it can only harness the salient information of one set of feature maps.
TAME also falls into the response-based category and operates in one forward pass, but contrarily to \cite{Gkartzonika2022}, it uses a trainable hierarchical attention module to exploit feature maps from multiple layers and generate explanations of higher quality.

We should also note that the methods of \cite{jetley2018learn, fukui2019attention} take a somewhat similar approach to ours in that they produce explanations using an attention module and multiple sets of feature maps.
However, these methods jointly train the attention model with the CNN to improve the image classification task.
In contrast, TAME does not modify the target model, which has been pretrained (and remains frozen); instead, TAME functions as a post-hoc method, exclusively optimizing the attention module in a supervised learning manner to generate visual explanations.
Thus, no direct comparisons can be drawn with \cite{jetley2018learn, fukui2019attention} as they provide explanations for a different (i.e. not the initial pretrained one), concurrently trained classifier.

\section{TAME}
\label{sec:tame}

\begin{figure*}[ht!]
\centering
\includegraphics[width=0.9\textwidth]{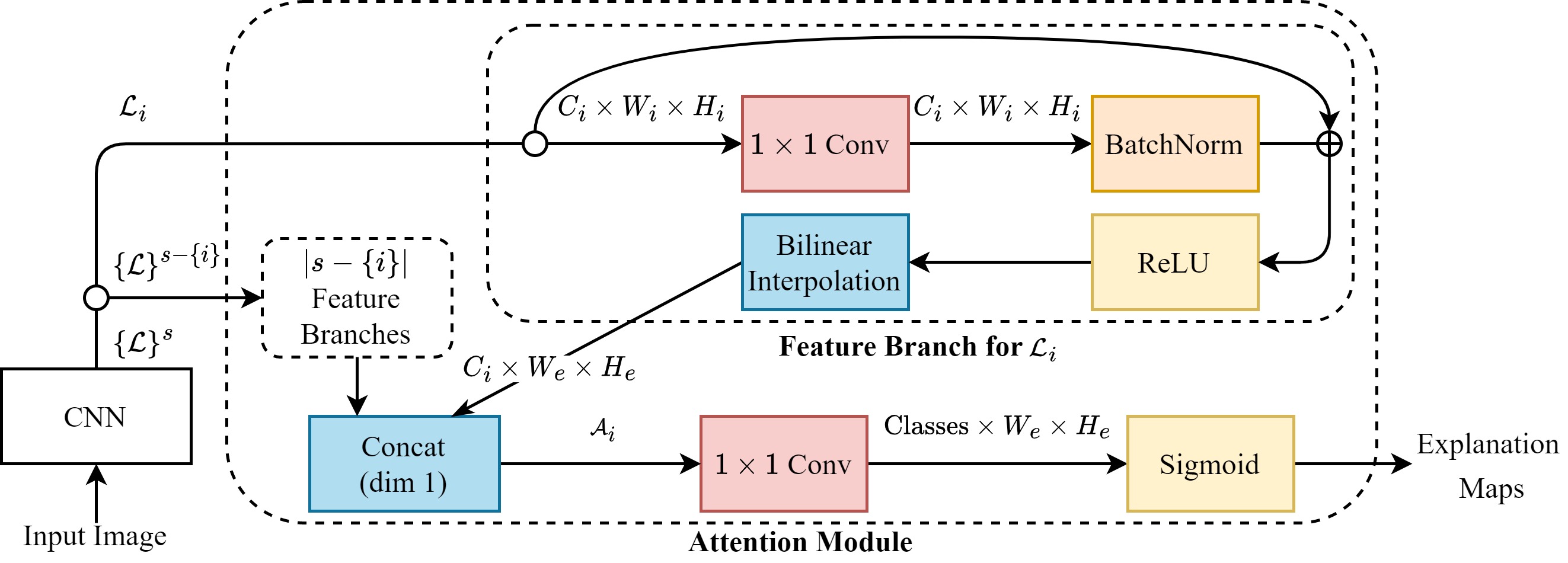} 
\caption{TAME's attention module: Feature branches process feature maps to provide attention maps, which are concatenated and processed by the fusion branch (shown at the bottom of the attention module) to derive explanation maps.}
\label{fig:arch}
\end{figure*}

\begin{figure}[ht!]%
\centering
\includegraphics[width=3.5in]{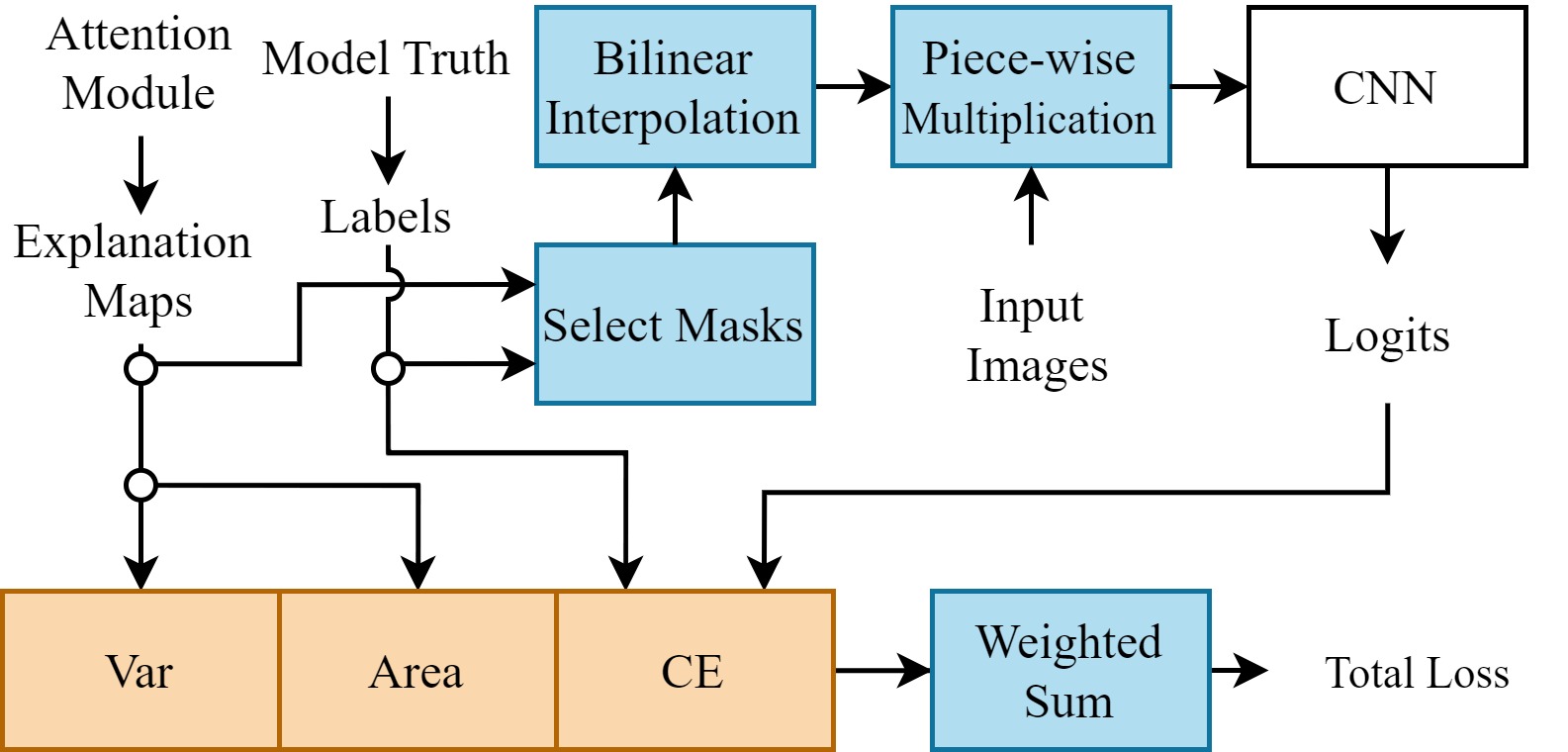} 
\caption{TAME's training method.
Var: Variation loss, Area: Area loss, CE: Cross-entropy loss.
The explanation of an input image is first derived; it is then upscaled and piece-wise multiplied with the corresponding input image.
Subsequently, the masked image does a second forward pass through the CNN to generate logits, which are used by the loss function to compute gradients and update the attention module's weights.}
\label{train}
\end{figure}%

\subsection{Problem formulation}
\label{ss:problemFormulations}

Let $f$ be a trained CNN for which we want to generate explanation maps,
\begin{equation}
    f: \mathcal{I} \to \mathbb{R}^{Classes},
\end{equation}
where, $\mathcal{I}$ is the set of all possible input tensors $\mathcal{I} = \left\{ \boldsymbol{I} \mid \boldsymbol{I} : \boldsymbol{C} \times \boldsymbol{W} \times \boldsymbol{H} \to \mathbb{R} \right\}$, $\boldsymbol{C} = \{1, \ldots, C\}$, $\boldsymbol{W} = \{1, \ldots, W\}$, $\boldsymbol{H} = \{1, \ldots, H\}$, $C, W, H \in \mathbb{R}$ are the input tensor dimensions \cite{sattarzadeh2021explaining,petsiuk2018rise}, and $Classes$ is the number of classes that $f$ has been trained to recognize.
E.g., for RGB images, $C=3$, and the elements of a tensor instance are the image pixel values.
Moreover, let $\mathcal{L}_i : 
\boldsymbol{C_i} \times \boldsymbol{W_i} \times \boldsymbol{H_i} \to \mathbb{R}$ be the feature map set corresponding to the $i$th layer of the CNN, where, $C_i, W_i, H_i$ are the respective channel, width and height dimensions.
We define a feature map set $\left\{\mathcal{L}\right\}^s$, where $s$ is the set of layers for which we want to extract feature maps, i.e., $\left\{\mathcal{L}\right\}^s = \{ \mathcal{L}_i | \; i \in s \}$.

Assume an attention module defined as in the following,
\begin{equation}
\text{AM}: \left\{\mathcal{L}\right\}^s \to \boldsymbol{E},    
\end{equation}
where, the tensor $\boldsymbol{E}$ at the output of the attention module is the generated explanation map,
$\boldsymbol{E} : \textbf{Classes} \times \boldsymbol{W_e} \times \boldsymbol{H_e} \to \left\{x \mid x \in \mathbb{R}\cap 0 < x < 1\right\}$, $W_e = \max \left\{W\right\}^s$ and $H_e = \max \left\{H\right\}^s$.
Thus, explanation maps are class discriminative, i.e., each slice of $\boldsymbol{E}$ along its first dimension corresponds to one of the classes that $f$ has learned; moreover, the size of the spatial dimensions of these ``class-specific'' slices equal to the largest spatial dimensions in the set of feature maps.

Given the above formulation, the goal is to find an attention module architecture that can combine all the salient information contained in $\left\{\mathcal{L}\right\}^s$, and effectively train it.

\subsection{Architecture}

We propose the attention module architecture depicted in Fig.~\ref{fig:arch}.
In this architecture, there exists a separate feature branch for each feature map set that is included in $\left\{\mathcal{L}\right\}^s$ and one fusion branch.
Each feature branch takes as input a single feature map set $\mathcal{L}_i$ and outputs an attention map set $\mathcal{A}_i$,
\begin{equation}
    \text{FB}: \mathcal{L}_i \to \mathcal{A}_i,
\end{equation}
where, $\mathcal{A}_i$ has the same channel and spatial dimensions as $\mathcal{L}_i$ and the final explanation map, respectively, i.e., $\mathcal{A}_i : \boldsymbol{C_i} \times \boldsymbol{W_e} \times \boldsymbol{H_e} \to \mathbb{R}$.
The resulting attention maps are concatenated into an single attention map set $\left\{\mathcal{A}\right\}^s$, and forwarded into the fusion branch to generate the explanation map,
\begin{equation}
    \text{FS}: \left\{\mathcal{A}\right\}^s \to \boldsymbol{E}.
\end{equation}
The two branch types consist of different network components, as described in the following:

\textit{Feature branch:} Each feature branch is a neural network that prepares the feature maps for the fusion branch.
It consists of a $1\times 1$ convolution layer with the same number of input and output channels, a batch normalization layer, a skip connection, a ReLU activation, and a bilinear interpolation that upscales the feature map to match the final explanation map's dimensions (the ablation study presented in Section \ref{sec:exp} assesses the importance of each part of the feature branch).
    
\textit{Fusion branch:} It consists of a $1\times 1$ convolutional layer that brings the number of the inputted channels to the number of classes that the CNN has been trained to recognize.
Subsequently, a sigmoid activation function, $S(x) = \frac{1}{1-e^{-x}}$, is used to scale the attention map values to the range $(0, 1)$.

\subsection{Training}

The training procedure is shown in Fig.~\ref{train}.
An image is inputted to the CNN model, and the derived feature maps are forwarded to the attention module for generating the respective explanation maps and the model truth label.
A model truth label is the CNN model's prediction of the input image's class, which may be different from the ground truth label.
A single channel containing a class discriminative \emph{explanation} is selected from the explanation map using the model truth label; this is used as the explanation of the input image with respect to the model truth class.
The explanation is then upscaled to the dimensions of the input image using bilinear interpolation, and is piece-wise multiplied with the input image.
The resulting masked image is then 
fed back into the CNN to generate logits.
The logits, the original explanation maps, and the model truth labels are then used to compute the loss and through backpropagation update the weights of the attention module, effectively training it.
As already mentioned, the weights of the original CNN remain fixed to their original values for the whole training procedure.

The loss function used for training the proposed attention module is the weighted sum of three individual loss functions,
\begin{eqnarray}
    L(\boldsymbol{\Psi}, \mbox{logits}, \mbox{labels}) &=& \lambda_1CE(\mbox{logits}, \mbox{labels}) \nonumber \\
    && + \, \lambda_2\mbox{Area}(\boldsymbol{\Psi}) + \lambda_3\mbox{Var}(\boldsymbol{\Psi}), \label{eq:loss}
\end{eqnarray}
where, $\boldsymbol{\Psi}$ is the slice of the explanation map $\boldsymbol{E}$ corresponding to the model truth class of the input image, $CE()$, Area(), Var() are the cross-entropy, area and variation loss, respectively, and $\lambda_1$, $\lambda_2$, $\lambda_3$, are the corresponding regularization parameters.
The cross-entropy loss uses the logits generated from the CNN with the masked input image and the model truth label to compute a loss value. This term trains the attention module to focus on salient parts of the image.
The variation loss is the sum of the squares of the partial derivatives of the explanation $\boldsymbol{\Psi}$ in the x and y direction.
This term penalizes fragmentation in the generated heatmaps.
For the partial derivatives, we use the forward difference approximation.
To this end, in the x direction we have $\frac{\partial \boldsymbol{\Psi}[x_m, y_m]}{\partial x} \approx \boldsymbol{\Psi}[x_m + 1, y_m] - \boldsymbol{\Psi}[x_m, y_m]$.
Thus, using the forward difference approximation the variation loss is defined as,
\begin{equation}
\text{Var}(\boldsymbol{\Psi}) = \sum_{x, y} \left[\left(\frac{\partial \boldsymbol{\Psi}[x, y]}{\partial x}\right)^2 + \left(\frac{\partial \boldsymbol{\Psi}[x, y]}{\partial y}\right)^2 \right].
\end{equation}
 Finally, the area loss is the mean of the explanation map $\boldsymbol{E}$ to the Hadamard power of $\lambda_4$, i.e.:
\begin{equation}
\label{eq:arealoss}
\text{Area}(\boldsymbol{\Psi}) = \sum_{x, y} {\boldsymbol{\Psi}[x, y]}^{\lambda_4}.
\end{equation}
This term forces the attention module to output heatmaps that emphasize small focused regions in the input image instead of arbitrarily large areas.

\subsection{Inference}

During inference, the final sigmoid activation function in the attention module (Fig. \ref{fig:arch}) is replaced with a min-max normalization operator, $m(x) = \frac{x - \min(\boldsymbol{\Psi})}{\max(\boldsymbol{\Psi}) - \min(\boldsymbol{\Psi})}$; the $\min()$ and $\max()$ operators return the smallest and largest element of the input tensor, respectively.
This is done for consistency with other literature works, such as \cite{selvaraju2017grad, chattopadhay2018grad}, on how the final explanation maps are scaled in order to be evaluated.
The test image is then forward-passed through the CNN, producing explanation maps, which are then upscaled to the input image size.
The derived model truth label can then be used to provide an explanation concerning the decision of the classifier.

\section{Experiments}\label{sec:exp}

\begin{figure*}[ht!]
\centering
\includegraphics[width=0.8\textwidth]{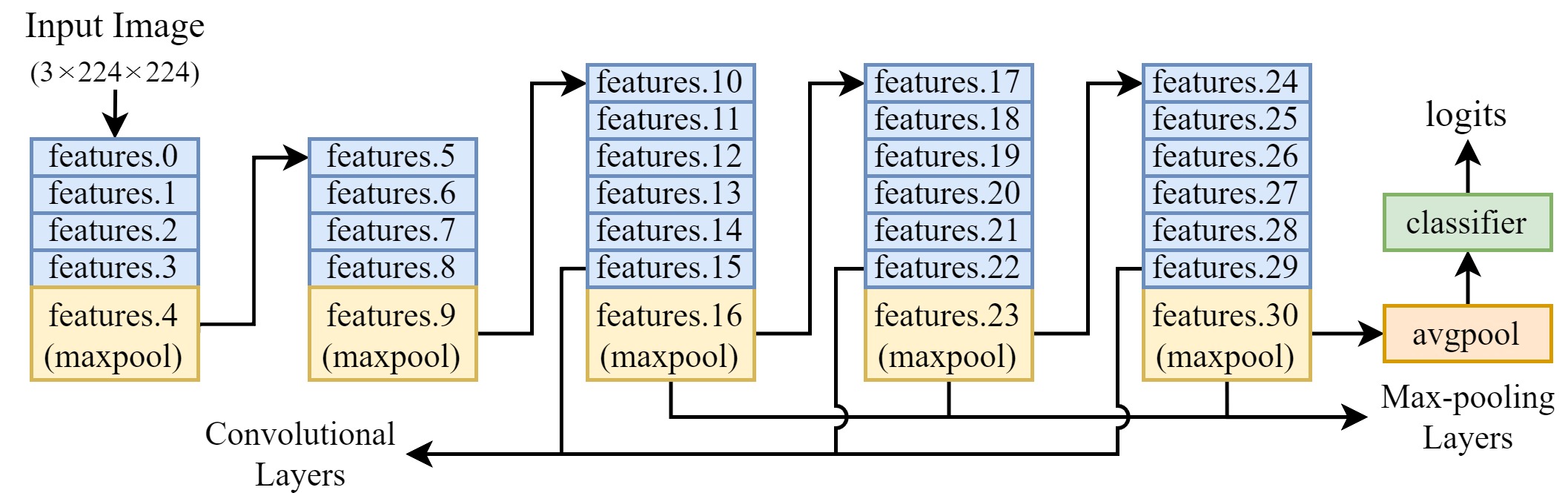} 
\caption{The layers from which feature map sets are extracted on VGG-16. We denote by ``Convolutional Layers'' the three layers before the last three max-pooling layers. In the case of VGG-16, the layer before a max-pooling layer is the ReLU activation function. We use the same layer naming as the library \texttt{torchvision.models.feature\_extraction}.}
\label{fig:vgg}
\end{figure*}

\subsection{Datasets and CNNs}\label{ssec:dataset}

We evaluate TAME on two popular CNNs pretrained on ImageNet: VGG-16 \cite{simonyan2014very} and ResNet-50 \cite{he2016deep}. 
We choose these two models to test the generality of our method because there are significant differences in the VGG and ResNet architectures.
We obtain these pretrained networks using the \verb|torchvision.models| library.

We train the attention module of our method with the ImageNet ILSVRC 2012 dataset \cite{ImageNet}.
This dataset contains 1000 classes, 1.3 million and 50k images for training and  evaluation, respectively.
Due to the prohibitively high cost of executing the literature's perturbation-based approaches that we use in the experimental comparisons, we use only 2000 randomly-selected testing images for testing (the same as in \cite{Gkartzonika2022} to allow a fair comparison) and a different 2000 randomly selected images as a validation set.

\subsection{Evaluation measures}

In the experimental evaluation, two frequently used evaluation measures, Increase in Confidence (IC) and Average Drop (AD) \cite{chattopadhay2018grad}, are utilized,
\begin{align}
     \text{AD}(v) &= \sum_{i=1}^{\Upsilon}\frac{\max(0, f(\boldsymbol{I}_i) - f(\boldsymbol{I}_i\odot \phi_v(\boldsymbol{\Psi}_i)))}{
     \Upsilon f(\boldsymbol{I}_i)} 100, \\
     \text{IC}(v) &= \sum_{i=1}^{\Upsilon}\frac{\text{sign}(f(\boldsymbol{I}_i\odot \phi_v(\boldsymbol{\Psi}_i))>f(\boldsymbol{I}_i))}{\Upsilon} 100,
\end{align}
where, $ \phi_v()$ is a threshold function to select the $v\%$ higher-valued pixels of the explanation map, $\text{sign}()$ returns 1 when the input condition is satisfied and 0 otherwise, $\Upsilon$ is the number of test images, $\boldsymbol{I}_i$ is the $i$th test image and $\boldsymbol{\Psi}_i$ is the corresponding explanation produced by TAME or any other method under evaluation. Intuitively, AD measures how much, on average, the produced explanation maps, when used to mask the input images, reduce the confidence of the model. In contrast, IC measures how often the explanation masks, when used to mask the input images, increase the confidence of the model. We threshold the explanation maps to test how well the pixels of the explanation map are ordered based on importance. Thus, using a smaller threshold results in a much more challenging evaluation setup.



\subsection{Experimental setup}

TAME is applied to VGG-16 using feature maps from three different layers.
The VGG-16 consists of five blocks of convolutions separated by $2\times 2$ max-pooling operations, as shown in Fig. \ref{fig:vgg}.
We choose one layer from each of the last three blocks, namely the feature maps output by the max-pooling layers of each block. We have also experimented on the feature maps output by the last convolution layer of each block.
On the other hand, ResNet-50 consists of five stages.
In the experimental evaluation, we use the feature maps from the final three stages.

TAME is trained using the loss function defined in \eqref{eq:loss} with the SGD (Stochastic Gradient Descent) algorithm.
The biggest batch size that can fit in the graphics card's memory is used, as recommended in \cite{smith2018disciplined}. 
The learning rate is varied using the OneCycleLR policy described in \cite{smith2019super}.
The maximum learning rate used by the OneCycleLR policy is chosen using the LR finder test defined in \cite{smith2017cyclical}.
The hyperparameters of the loss function (\eqref{eq:loss}, \eqref{eq:arealoss}) are empirically chosen using the validation dataset, as: $\lambda_1 = 1.5, \lambda_2 = 2, \lambda_3 = 0.01, \lambda_4 = 0.3$.

We train the attention module for eight epochs in total and select the epoch for which the attention module achieved the best $\text{IC}(15\%)$ and $\text{AD}(15\%)$ in the validation set.
That is, in this model selection procedure we opt for the measures at the $15\%$ threshold because they are the most challenging measures to improve upon and provide more focused explanation masks.

During training, each image is transformed in the same way as with the original CNN, i.e., its smaller spatial dimension is resized to 256 pixels, random-cropped to dimensions $W = H = 224$, and normalized to zero mean and unit variance.
The same is done during testing, except that center-cropping is used. The feature maps are extracted from the CNN using \verb|torchvision.models.feature_extraction| library.

\begin{table*}[th!]
\caption{Comparison of TAME with other methods}
\label{tab:comp}
\centering
\begin{tabular}{cccccccc}
\toprule
Model & Measure & Grad-CAM \cite{selvaraju2017grad}& Grad-CAM++ \cite{chattopadhay2018grad}& Score-CAM \cite{wang2020score}& RISE \cite{petsiuk2018rise}& L-CAM-Img \cite{Gkartzonika2022}& TAME \\
\midrule
\multirow{7}{*}{VGG-16}
&$\text{AD}(100\%)$ & 32.12 & 30.75 & 27.75 & \textbf{8.74} & 12.15 & \underline{9.33} \\
&$\text{IC}(100\%)$ & 22.1 & 22.05 & 22.8 & \textbf{51.3} & 40.95 & \underline{50} \\
\cmidrule{2-8}
&$\text{AD}(50\%)$ & 58.65 & 54.11 & 45.6 & 42.42 & \underline{37.37} & \textbf{36.5} \\
&$\text{IC}(50\%)$ & 9.5 & 11.15 & 14.1 & 17.55 & \underline{20.25} & \textbf{22.45}\\
\cmidrule{2-8}
&$\text{AD}(15\%)$ & 84.15 & 82.72 & 75.7 & 78.7 & \underline{74.23} & \textbf{73.29}\\
&$\text{IC}(15\%)$ & 2.2 & 3.15 & 4.3 & \underline{4.45} & \underline{4.45} & \textbf{5.6} \\
\cmidrule{2-8}
& Forward Passes (Inference) & \textbf{1} & \textbf{1} & 512 & 4000 & \textbf{1} & \textbf{1} \\
\midrule
\multirow{7}{*}{ResNet-50}
&$\text{AD}(100\%)$ & 13.61 & 13.63 & \underline{11.01} & 11.12 & 11.09 & \textbf{7.81} \\
&$\text{IC}(100\%)$ & 38.1 & 37.95 & 39.55 & \underline{46.15} & 43.75 & \textbf{54} \\
\cmidrule{2-8}
&$\text{AD}(50\%)$ & 29.28 & 30.37 & \textbf{26.8} & 36.31 & 29.12 & \underline{27.88} \\
&$\text{IC}(50\%)$ & 23.05 & 23.45 & \underline{24.75} & 21.55 & 24.1 & \textbf{27.5} \\
\cmidrule{2-8}
&$\text{AD}(15\%)$ & \underline{78.61} & 79.58 & 78.72 & 82.05 & 79.41 & \textbf{78.58} \\
&$\text{IC}(15\%)$ & 3.4 & 3.4 & 3.6 & 3.2 & \underline{3.9} & \textbf{4.9} \\
\cmidrule{2-8}
& Forward Passes (Inference) & \textbf{1} & \textbf{1} & 2048 & 8000 & \textbf{1} & \textbf{1} \\
\bottomrule
\end{tabular}
\end{table*}

\begin{table*}[ht!]
\caption{Ablation study of TAME}
\label{tab:abl}
\centering
\begin{tabular}{ccccccccc}
\toprule
Model & Feature Extraction & Architecture Variant & $\text{AD}(100\%)$ & $\text{IC}(100\%)$ & $\text{AD}(50\%)$ & $\text{IC}(50\%)$ & $\text{AD}(15\%)$ & $\text{IC}(15\%)$ \\
\midrule
\multirow{12}{*}{VGG-16} 
& \multirow{6}{*}{Max-pooling layers}
&Proposed Architecture & 9.33 & 50 & 36.5 & 22.45 & \underline{73.29} & \underline{5.6} \\
&&No skip connection & 10.09 & 45.25 & 36.44 & 20.65 & 74.85 & 5.15 \\
&&No skip + No batch norm & \textbf{5.92} & \textbf {57.9} & \underline{34.49} & \textbf{24.2} & 74.58 & 5.15 \\
&&Sigmoid in feature branch & \underline{7.22} & \underline{55.65} & 38.4 & 21.6 & 79 & 4.85 \\
&&Two layers & 10.72 & 45.45 & \textbf{34.48} & \underline{23.05} & \textbf{71.94} & \textbf{5.75} \\
&&One layer & 12.1 & 42.1 & 35.81 & 20.8 & 74.19 & 4.85 \\
\cmidrule{2-9}
&\multirow{6}{*}{Convolutional layers}
&Proposed Architecture & 9.07 & 51.1 & 40.72 & \textbf{20.9} & \underline{77.05} & \underline{4.8} \\
&&No skip connection & \textbf{6.22} & \underline{58.85} & 41.47 & \textbf{20.9} & 79.12 & 3.8 \\
&&No skip + No batch norm & \underline{6.62} & 56.6 & \textbf{40.48} & \underline{20.75} & 77.84 & \textbf{4.95} \\
&&Sigmoid in feature branch & 6.8 & \textbf{60} & 42.17 & 19.75 & 80.73 & 4.1 \\
&&Two layers & 10.99 & 45.85 & \underline{40.89} & 19.55 & \textbf{76.66} & \underline{4.8} \\
&&One layer & 13.09 & 39.65 & 42.3 & 17.7 & 78.02 & 3.8 \\
\midrule
\multirow{6}{*}{ResNet-50} 
&\multirow{6}{*}{Stage Outputs}
&Proposed Architecture & \underline{7.81} & \underline{54} & \underline{27.88} & \textbf{27.5} & \underline{78.58} & \textbf{4.9} \\
&&No skip connection & \textbf{5.7} & \textbf{62.65} & 46.58 & 18.25 & 89.32 & 2.3 \\
&&No skip + No batch norm & 9.29 & 50.25 & 29.43 & \underline{25.95} & 79.81 & 3.95 \\
&&Sigmoid in feature branch & 9.11 & 53.3 & 45.68 & 18.1 & 86.95 & 3.15 \\
&&Two layers & 9.48 & 47.05 & \textbf{27.83} & 25 & \textbf{77.95} & \underline{4.25} \\
&&One layer & 11.32 & 43.45 & 29.85 & 24.25 & 79.59 & 3.55 \\
\bottomrule

\end{tabular}
\end{table*}

\subsection{Quantitative Evaluation}

The proposed method is compared against the top-performing approaches in the visual explanation domain for which source code is publicly available i.e. Grad-CAM \cite{selvaraju2017grad}, Grad-CAM++ \cite{chattopadhay2018grad}, Score-CAM \cite{wang2020score}, RISE \cite{petsiuk2018rise} and L-CAM \cite{Gkartzonika2022}.
The performance is measured using $\text{AD}(v)$ and $\text{IC}(v)$ on three different thresholds $v$ of increasing difficulty, i.e., $v = 100\%, 50\%$ and $15\%$, similarly to the evaluation protocol of \cite{Gkartzonika2022}.
An ablation study is also conducted, to assess the importance of the different architecture components for VGG-16 and ResNet-50, as well as to showcase the effect of different layer selections in the VGG-16 model.

\subsubsection{Comparison with the State-Of-The-Art}
In Table~\ref{tab:comp} we highlight with bold letters the best result and underline the second best result for each measure, separately for each base model. We can see that TAME outperforms the gradient-based methods, and is competitive to the perturbation-based methods, obtaining the best results for the more demanding $15\%$ measures while requiring only one forward pass.

\subsubsection{Ablation Study} \label{sssec:abl}

In Table~\ref{tab:abl} we highlight with bold letters the best results and underline the second best result for each measure in each model and layer selection.
For the VGG-16 model, inspired from similar works in the literature suggesting that the last layers of the network provide more salient features \cite{jetley2018learn}, we report two sets of experiments, one that uses features extracted from the three last max-pooling layers and one where features are extracted from the layers exactly before the last three max-pooling layers (Fig.~\ref{fig:vgg}). 
There is a difference in the spatial dimensions of the explanation maps generated using the former or the latter layers for feature extraction, i.e., $28 \times 28$ versus $56\times 56$, since the dimension of the explanation maps obtained by TAME is dictated by the largest feature map set (as explained in Section \ref{sec:tame}). For the ResNet-50 model, we extract features from the outputs of the final three stages, resulting to an explanation map of $28 \times 28$ spatial dimensions. We examine the following variants of the proposed architecture:

\textit{No skip connection}: It has been shown that the skip connection promotes a smoother loss landscape \cite{li2018visualizing}, thus contributing to training very deep neural networks. Even for shallower neural networks, such as the proposed attention module, we can benefit from using a skip connection. We see that by omitting the skip connection, we get worse results in ResNet-50. Similarly, for both baseline models we report worse performance for the harder $50\%$ and $15\%$ measures.

\begin{figure}[!th]
\centering
\includegraphics[width=3.5in]{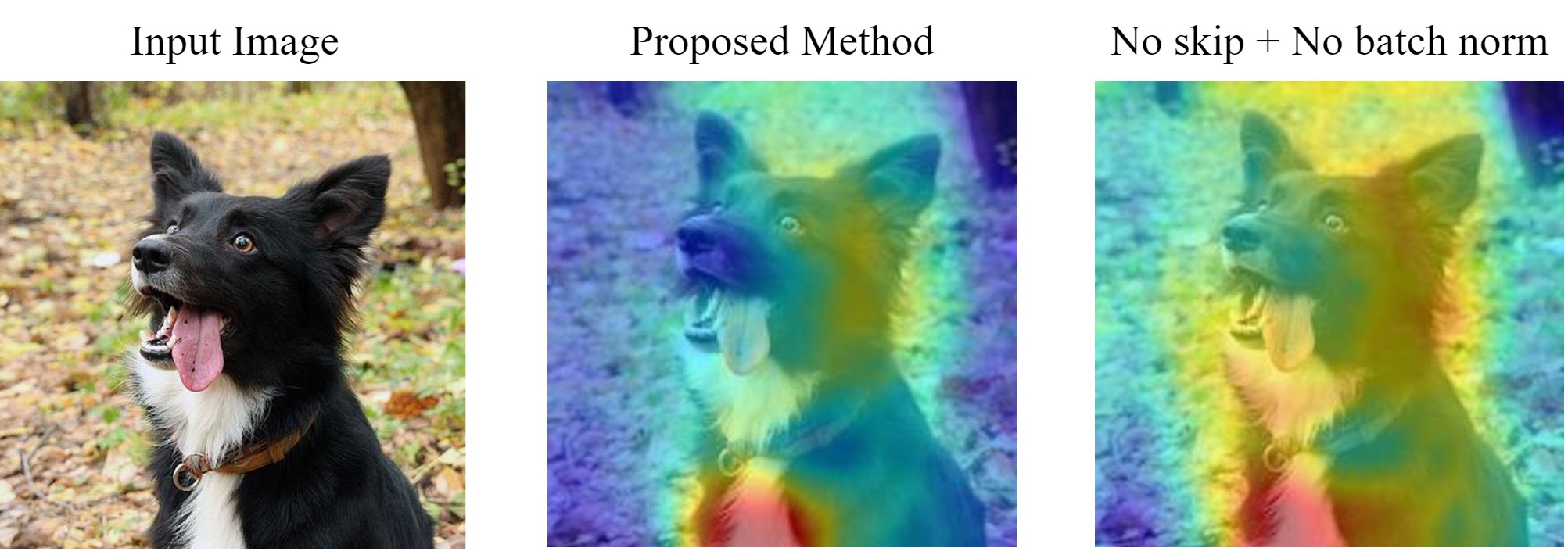}
\caption{Qualitative comparison between the proposed attention module and the `no skip + no batch norm' variant, applied to VGG-16.
We observe that for the `no skip + no batch norm' architecture, the produced explanation map is more spread out, showing that even if it performs well on the $100\%$ measures, it fails to precisely identify the salient regions in the image.
}
\label{fig:comp}
\end{figure}

\textit{No skip + No batch norm}: Batch normalization is used in CNNs for speeding up training and combating internal covariate shift \cite{ioffe2015batch}.
Compared to the proposed architecture, we see that this variant generally performs better in the $100\%$ measures, but this does not hold for the other measures.
We compare the masks produced by this variant in Fig.~\ref{fig:comp}.

\textit{Sigmoid in feature branch}: In this variant we replace the ReLU function with the sigmoid function, which squeezes the input from $(-\infty, \infty)$ to the output $(0, 1)$.
It is well known that the sigmoid function in deeper neural networks causes the vanishing gradient problem, making it more difficult to train the early layers of the CNN. We see again that the proposed architecture prevails for the more challenging $15\%$ measures.

\textit{Two layers} and \textit{One layer}: In this case, the proposed attention module architecture is employed with fewer feature maps.
The results when using just one layer, i.e., omitting the two earlier layers in the CNN pipeline (Fig.~\ref{fig:vgg}), are very similar to the L-CAM-Img method (as shown in Table~\ref{tab:comp}), which also uses just one feature map set.
All measures are improved when utilizing a second feature map set, i.e., excluding only the earliest layer in the CNN pipeline; however, the case is not the same clear when going from the two to three feature map sets, which are used in the proposed architecture.
These mixed results could be attributed to the extra noise of feature maps taken earlier in a CNN pipeline.

We note that by omitting both the skip connection and the batch normalization in the feature branch architecture, we obtain generally better results in the case of the VGG-16 model, but this is not the case for the same architecture applied to the ResNet-50 model.
In addition, all architectures struggle with the more difficult $15\%$ measures compared to the proposed architecture.
Although every architecture varies between models, the proposed architecture generalizes best across different models.
Thus, our goal of finding an effective architecture across radically different models is achieved through the proposed architecture.

\begin{figure}[!ht]
\centering
\includegraphics[width=3.5in]{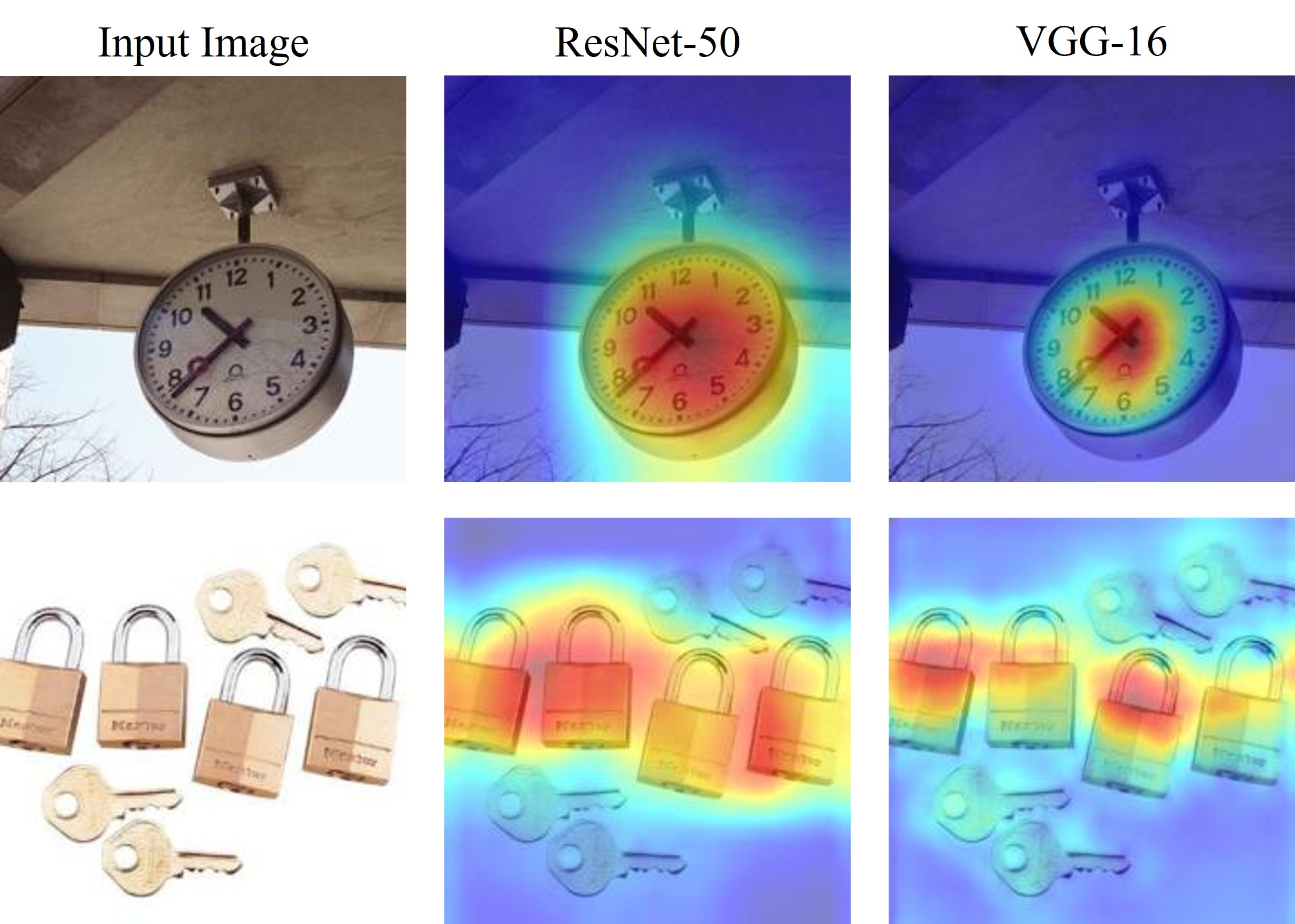}
\caption{TAME applied to ResNet-50 and VGG-16 for the ground truth class (Top image: ``analog clock'' (406), Bottom image: ``padlock'' (695).
The explanation masks produced using the VGG-16 are more focused in comparison to the ones of ResNet-50.
}
\label{fig:fig5}
\end{figure}

\begin{figure*}[!ht]
    \centering
    \adjustbox{minipage=2em,valign=t}{\subcaption{}\label{fig:fig6a}}%
    \begin{subfigure}{\dimexpr\linewidth-2em\relax}
        \centering%
        \includegraphics[width=\linewidth]{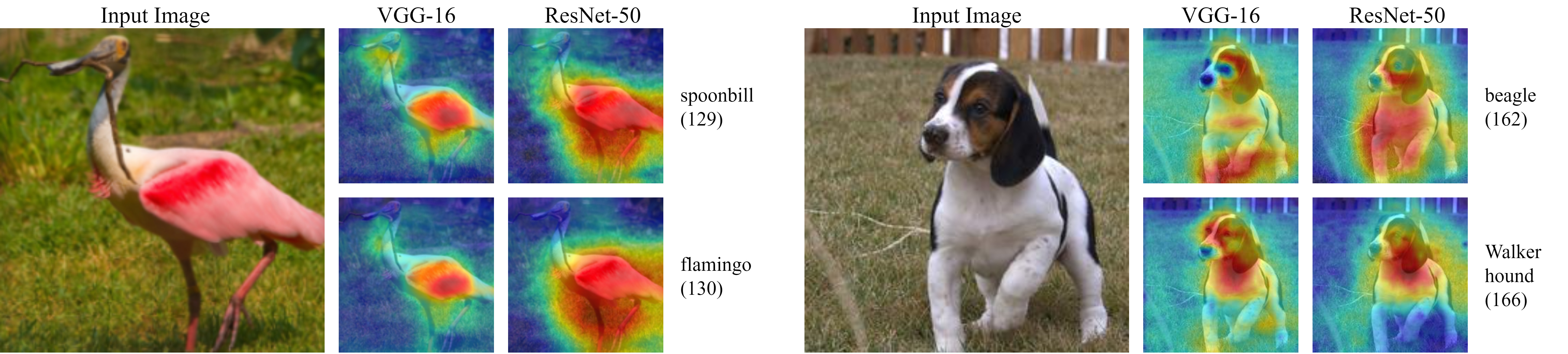}
        \vspace{\dimexpr\linewidth / 120\relax}%
    \end{subfigure}\\%
    \adjustbox{minipage=2em,valign=t}{\subcaption{}\label{fig:fig6b}}%
    \begin{subfigure}{\dimexpr\linewidth-2em\relax}
        \centering%
        \includegraphics[width=\linewidth]{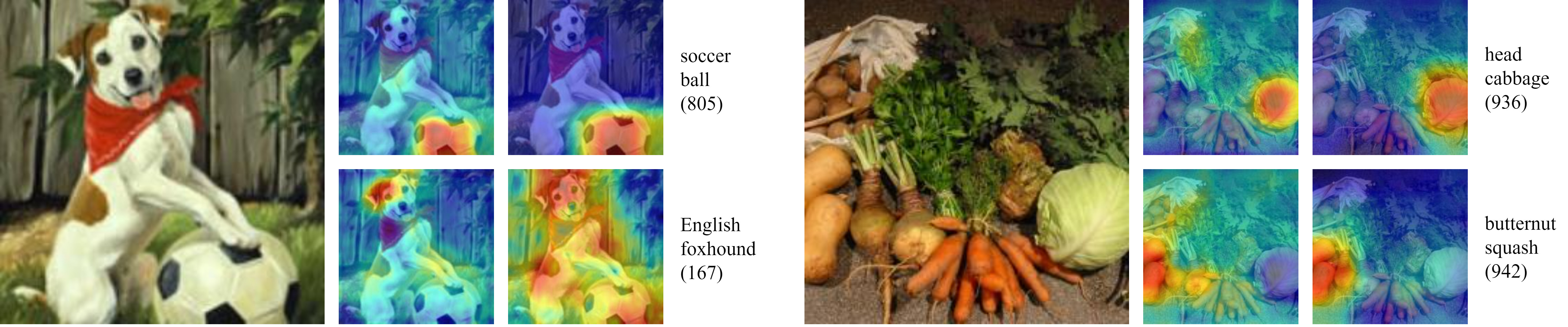}
        \vspace{\dimexpr\linewidth / 120\relax}%
    \end{subfigure}\\%
    \adjustbox{minipage=2em,valign=t}{\subcaption{}\label{fig:fig6c}}%
    \begin{subfigure}{\dimexpr\linewidth-2em\relax}
        \centering
        \includegraphics[width=\linewidth]{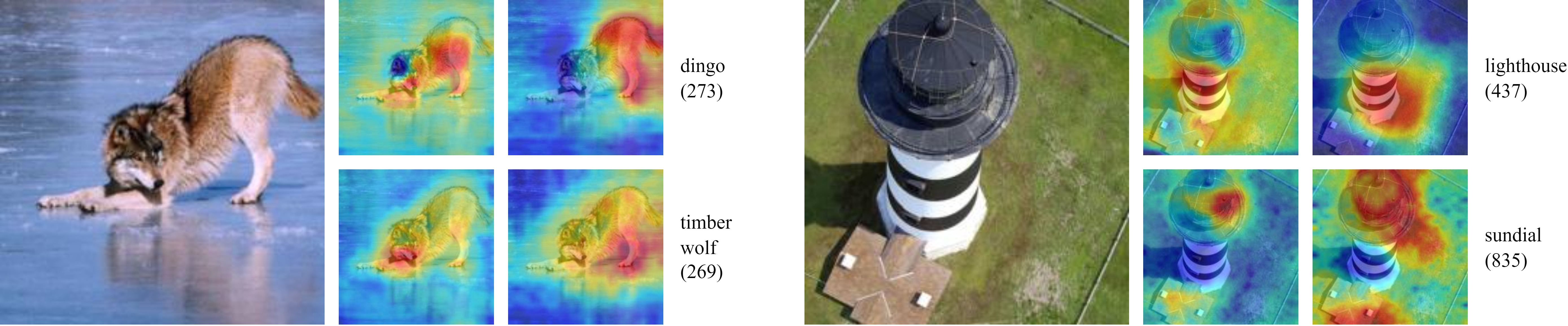} 
    \end{subfigure}
    \caption{Explanation for six input images.
    In each case we display four class-specific explanations, i.e., of the true (top) and an erroneous (bottom) class prediction of the input image, for both ResNet-50 and VGG-16.
    Figs.~\ref{fig:fig6a}, ~\ref{fig:fig6b} depict examples of images with similar classes and with images containing multiple classes, respectively.
    In Fig.~\ref{fig:fig6c} two cases of misclassification are provided: dataset misclassification (left side example) and model misclasssification (right side example). 
    }
    \label{fig:fig6}
\end{figure*}

\subsection{Qualitative Analysis}

An extensive qualitative analysis is also performed using the ILSVRC 2012 ImageNet dataset in order to gain insight of the proposed approach and appreciate its usefulness in real-world applications, e.g., understanding why an image was correctly classified or misclassified.
The examples used in this study are depicted in Figs.~\ref{fig:comp}, \ref{fig:fig5} and \ref{fig:fig6}.

Fig.~\ref{fig:comp} compares TAME generated explanation maps with explanations generated by the ``No skip + No batch norm'' architecture examined in Section \ref{sssec:abl}.
The improved ability of TAME to identify the salient image regions highlights the importance of evaluating the method using the $\text{AD}$ and $\text{IC}$ measures on multiple thresholds (Table~\ref{tab:abl}), and particularly the significance of the $15\%$ measures over the $100\%$ and $50\%$ ones in determining the quality of generated explanations.

The differences between explanations produced using TAME on ResNet-50 and VGG-16 are examined in Fig.~\ref{fig:fig5}.
We observe that explanations produced for the ResNet-50 model are generally more activated, and, in general, explanations produced for the two different CNN types attend different areas of the image.
This suggests that ResNet-50 and VGG-16 classify images in fundamentally different ways, focusing on different features of an input image to make their predictions.

In Fig. \ref{fig:fig6}, we provide class-specific explanation masks referring to the ground truth class but also to an erroneous but closely related class, for both ResNet-50 and VGG-16 models.
The first image of Fig.~\ref{fig:fig6a}, depicts a spoonbill, a bird similar to the flamingo.
Two significant differences between the spoonbill and the flamingo are the characteristic bill and the darker pink stripe on the wing of the spoonbill.
We can see in the explanation maps of both models, that when choosing the class flamingo, there is no significance attributed to the bill, but, on the other hand, when the spoonbill class is chosen, the bill area is gaining significant attention.
By comparing the explanation maps for adversarial classes, we can gain insight into important features which characterize a specific object against similar ones, and possibly gain new insight \emph{from} the classifier.
The second image in Fig.~\ref{fig:fig6a} is a similar case.

The examples of Fig.~\ref{fig:fig6b} demonstrate the potential of the explanation maps to be used for explaining multiple different classes contained in a single image, i.e., the ``english foxhound'' and ``soccer ball'' image, and the ``head cabbage'' and ``butternut squash'' image.

Finally, in Fig.~\ref{fig:fig6c} we provide two cases of images that have been miscategorized, and use the explanations to understand what has gone wrong.
The first image of Fig.~\ref{fig:fig6c} belongs to the ``dingo'' class (273) but is evidently misclassified as ``timber wolf'' from both CNN models. 
Using the explanations, we can identify important features on the image for each class and CNN model.
The second image depicts a lighthouse.
VGG-16 misclassifies this image as a ``sundial''.
Again, using the explanations generated by TAME we can understand which features led the model to produce a wrong decision.
For instance, in this case, we see that for both models the ``sundial'' explanations focus on the lighthouse roof, which might resemble a sundial, explaining the erroneous classification decision of VGG-16.

\section{Conclusions}

We proposed TAME, a novel method for generating visual explanations for various CNNs.
This is accomplished by training a hierarchical attention module to extract information from feature map sets of multiple layers.
Experimental results verified that TAME outperforms gradient-based methods and competes with perturbation-based, while, in contrast to them, requires only a single forward pass to generate explanations.
Further research is needed to discover the limits of the proposed approach, e.g., generalizing it to non-CNN architectures.





\end{document}